%% file: main.tex
\documentclass[lettersize,journal]{IEEEtran}
\usepackage{amsmath,amsfonts}
\usepackage{algorithmic}
\usepackage{array}
\usepackage[caption=false,font=normalsize,labelfont=sf,textfont=sf]{subfig}
\usepackage{textcomp}
\usepackage{stfloats}
\usepackage{url}
\usepackage{verbatim}
\usepackage{graphicx}
\usepackage{amsthm}
\usepackage{cite}
\usepackage{graphicx}
\usepackage{amssymb}
\usepackage[ruled,vlined]{algorithm2e}
\usepackage{multirow}
\usepackage{svg}

\usepackage{multicol}

\newcommand{\OURS}{GPFL}

\newtheorem{theorem}{Theorem}

\newtheorem{assumption}{Assumption}

\def\BibTeX{{\rm B\kern-.05em{\sc i\kern-.025em b}\kern-.08em
    T\kern-.1667em\lower.7ex\hbox{E}\kern-.125emX}}
\usepackage{balance}
\begin{document}
\title{GPFL: A Gradient Projection-Based Client Selection Framework for Efficient Federated Learning}


\author{Shijie Na, Yuzhi Liang*, Siu-Ming Yiu
\thanks{Shijie is with Computer Science Department, The University of Hong Kong, email: shijiena@connect.hku.hk}
\thanks{Yuzhi Liang is with Information Science and Technology, Guangdong University of Foregin Studies, email: yzliang@gdufs.edu.cn}
\thanks{Siu-Ming Yiu is with Computer Science Department, The University of Hong Kong, email:  smyiu@cs.hku.hk}
}

\markboth{Journal of \LaTeX\ Class Files,~Vol.~18, No.~9, September~2020}%
{How to Use the IEEEtran \LaTeX \ Templates}

\maketitle

\begin{abstract}
Federated learning client selection is crucial for determining participant clients while balancing model accuracy and communication efficiency. Existing methods have limitations in handling data heterogeneity, computational burdens, and independent client treatment. To address these challenges, we propose {\OURS}, which measures client value by comparing local and global descent directions. We also employ an Exploit-Explore mechanism to enhance performance. Experimental results on FEMINST and CIFAR-10 datasets demonstrate that {\OURS} outperforms baselines in Non-IID scenarios, achieving over 9\% improvement in FEMINST test accuracy. Moreover, {\OURS} exhibits shorter computation times through pre-selection and parameter reuse in federated learning.
\end{abstract}


\input{1introduction}

\input{2Motivation}

\input{3Design}
\input{4Evaluation}

\input{8Relatedwork}

\input{9Conclusion}

\input{10Appendix}
\bibliographystyle{IEEEtran}

\bibliography{reference}

 

\end{document}

%% file: 1introduction.tex
\section{Introduction}
Federated learning (FL) is a promising approach to collaborative model training that preserves privacy while leveraging the resources of multiple clients. However, in real-world scenarios, redundant updates from similar gradients can lead to inefficiencies. Therefore, the selection of clients for training is crucial to improve convergence rates, model accuracy, and fairness.

Existing client selection algorithms for FL have limitations in addressing challenges of data quality evaluation, resource consumption minimization, and accounting for client interdependence. In terms of data quality evaluation, previous approaches have relied on metrics such as the Shapley value and data size \cite{nagalapatti2021game}. However, computing the Shapley value is computationally expensive and impractical for scenarios with a large number of clients. Consequently, many algorithms have resorted to assessing data quality solely based on data size, disregarding the Non-IID nature of client data. This approach inaccurately measures data quality by considering quantity alone. In the context of resource consumption, many existing client selection algorithms employ a post-selection approach, which involves distributing the model to each client for local computations of gradients and other relevant information \cite{balakrishnan2022diverse}. However, this method incurs significant communication and computation costs. The parameter server must distribute the global model to each client, and each client carries an additional computational burden to enhance client selection accuracy. Regarding the third issue, lots of previous studies have overlooked the interdependence among clients in FL, treating each client as an independent contributor and aggregating their gradients \cite{le2021incentive}. However, the presence of dependencies or correlations between different clients can have a significant impact on the model accuracy.

Resolving the aforementioned issues in FL is a non-trivial task. FL requires strict adherence to user privacy protection, which makes it impossible to directly assess data quality with plaintext client data. On the other hand, accurately examining the interaction between clients necessitates calculating the effects of all possible combinations and selecting the most optimal one. However, when dealing with a large number of clients, the sheer magnitude of potential combinations becomes overwhelmingly vast, making it impractical to exhaustively explore all possibilities and identify the absolute best choice.

In this paper, we present a novel framework called \textbf{Gradient Projection-based Federated Learning ({\OURS})} for client selection in FL. 
The core concept of {\OURS} is to ascertain the precise direction of gradient descent by evaluating the proximity of each client's momentum-based gradient to the overall gradient direction. This methodology facilitates the selection of clients whose data align closely with the training objective, effectively reducing the influence of stochastic gradient noise and improving the overall smoothness of the learning process.

Moreover, {\OURS} incorporates a mechanism that combines exploitation and exploration to select the optimal combination of clients. As a pre-selection method, {\OURS} first identifies suitable clients based on their historical involvement in training. It then distributes the global model to the selected clients for local training, minimizing the computational burden on clients and reducing communication costs between clients and the server. In general, our scheme guarantees fast and precise client selection. We have conducted thorough theoretical and experimental analyses to validate the effectiveness of our approach. The main contributions of our work can be summarized as follows.

\begin{itemize}
\item We propose {\OURS}, a gradient projection-based client selection framework designed to enhance the efficiency of federated learning. As a pre-selection method, {\OURS} focuses on the latest gradient submission among selected clients, rather than waiting for submissions from all clients. This approach significantly reduces the waiting time for model updates. 
\item {\OURS} utilizes a novel indicator called \textbf{Gradient Projection (GP)}, enabling efficient and precise assessment of the quality of client data. Our proposed metric provides an accurate assessment of the impact of client data on the training process while also preserving client data privacy. By leveraging this indicator, {\OURS} efficiently selects clients with high-quality data for improved training outcomes. 
\item {\OURS} effectively addresses the Exploration-Exploitation trade-off in client selection for FL by combining the gradient projection metric with the Gradient Projection Confidence Bound function. This integration enables {\OURS} to intelligently explore the client space, considering the relationships between clients' data and achieving a balance between exploring new client combinations and exploiting the knowledge gained from previously selected clients. By leveraging the proposed Gradient Projection Confidence Bound, {\OURS} maximizes the utilization of available client resources while ensuring the selection of high-quality clients. It takes into account the potential benefits of unexplored clients while also considering the confidence in the performance of already selected clients. We theoretically demonstrated that the selected client set chosen by {\OURS} closely approximates the optimal solution, reinforcing the effectiveness and reliability of our approach. 
\end{itemize}

%% file: 2Motivation.tex
\section{Federated Learning with Client Selection}


In a typical federated learning (FL) scenario, the client selection process begins with the server distributing its initial model parameters, denoted as $w^0$, to each participating client. The clients then compute an initial gradient based on their local data and send it back to the global model. This process is typically repeated in subsequent rounds until convergence is achieved. In each training round, the following steps are performed:

\begin{enumerate}
\item \textbf{Client selection:} There are two main approaches to client selection in FL, namely pre-selection methods like our designed method and FedCor \cite{tang2022fedcor}, and post-selection methods like DivFL \cite{balakrishnan2022diverse}. In the pre-selection mechanism, the server first identifies suitable clients based on information from the previous round and then distributes the updated global model to those selected clients. Each of these clients computes the gradient using their local data and uploads it to the server. On the other hand, in the post-selection mechanism, the server initially distributes the parameters of the global model to all clients in round t. All clients then compute gradients based on their local data and upload them to the server. The server selects suitable candidates to participate in FL training based on the submitted gradients.

\item \textbf{Model aggregation:} The server performs aggregation of the information submitted by clients and calculates the updated global model parameters. One commonly used model aggregation strategy is FedAvg \cite{DBLP:conf/iclr/LiHYWZ20}. In FedAvg, during round t, client i calculates a new local model $w_i^{t}$ using the global model $w^{t-1}$ and its local data. Subsequently, the weight of the global model is updated as follows:
\[ w^{t} = \frac{1}{|\mathcal{S}_t|}\sum_{i=1}^{|\mathcal{S}_t|} w_i^{t} \]
where $\mathcal{S}_t$ denotes the set of selected clients in round t.


\item \textbf{Performance Evaluation:} The performance of the global model is evaluated by assessing various metrics such as accuracy, loss, and other relevant indicators. This evaluation step serves to gauge the effectiveness of the selected clients and the overall learning process.
\end{enumerate}

%% file: 3Design.tex
\section{Methodology}
A key step of client selection is to assess the quality of the client's data and choose those clients that have the most significant contribution to the training. This section presents an efficient method for client selection that facilitates rapid evaluation of client's data and guarantees high performance and convergence rate in the trained model. More precisely, we have developed a gradient projection technique to evaluate the data quality of each client, followed by aggregating the momentum-based gradients of the selected clients to determine the gradient direction.

\subsection{Gradient Projection}
\label{subsec: MGP}
\begin{figure*}[ht]
  \centering
  \begin{minipage}[b]{0.45\linewidth}
    \centering
\includegraphics[width=\linewidth]{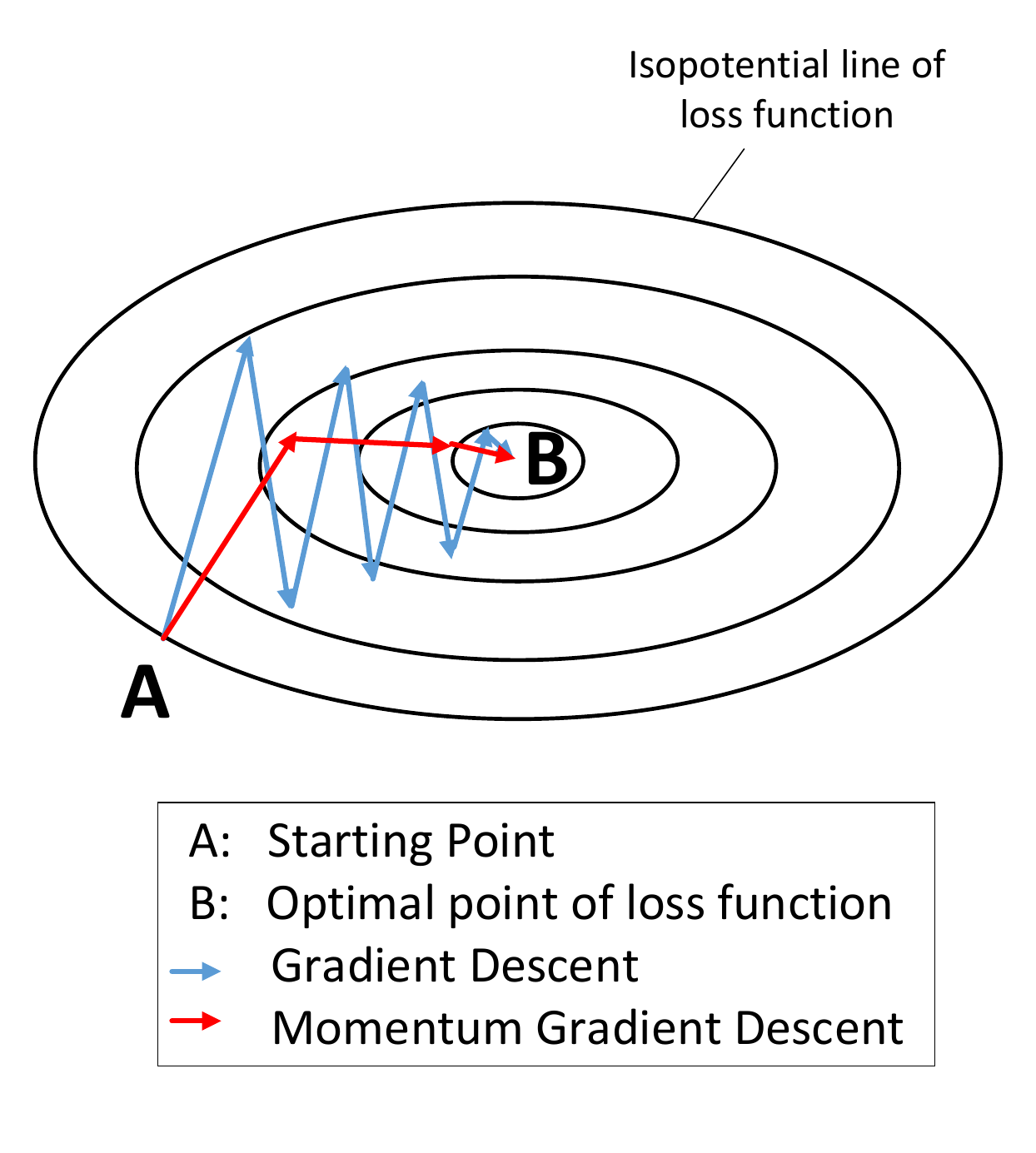}
  \caption{MGD vs GD}
  \label{fig:MGDvsGD}
  \end{minipage}
  \hfill
  \begin{minipage}[b]{0.45\linewidth}
    \centering
    \includegraphics[width=\linewidth]{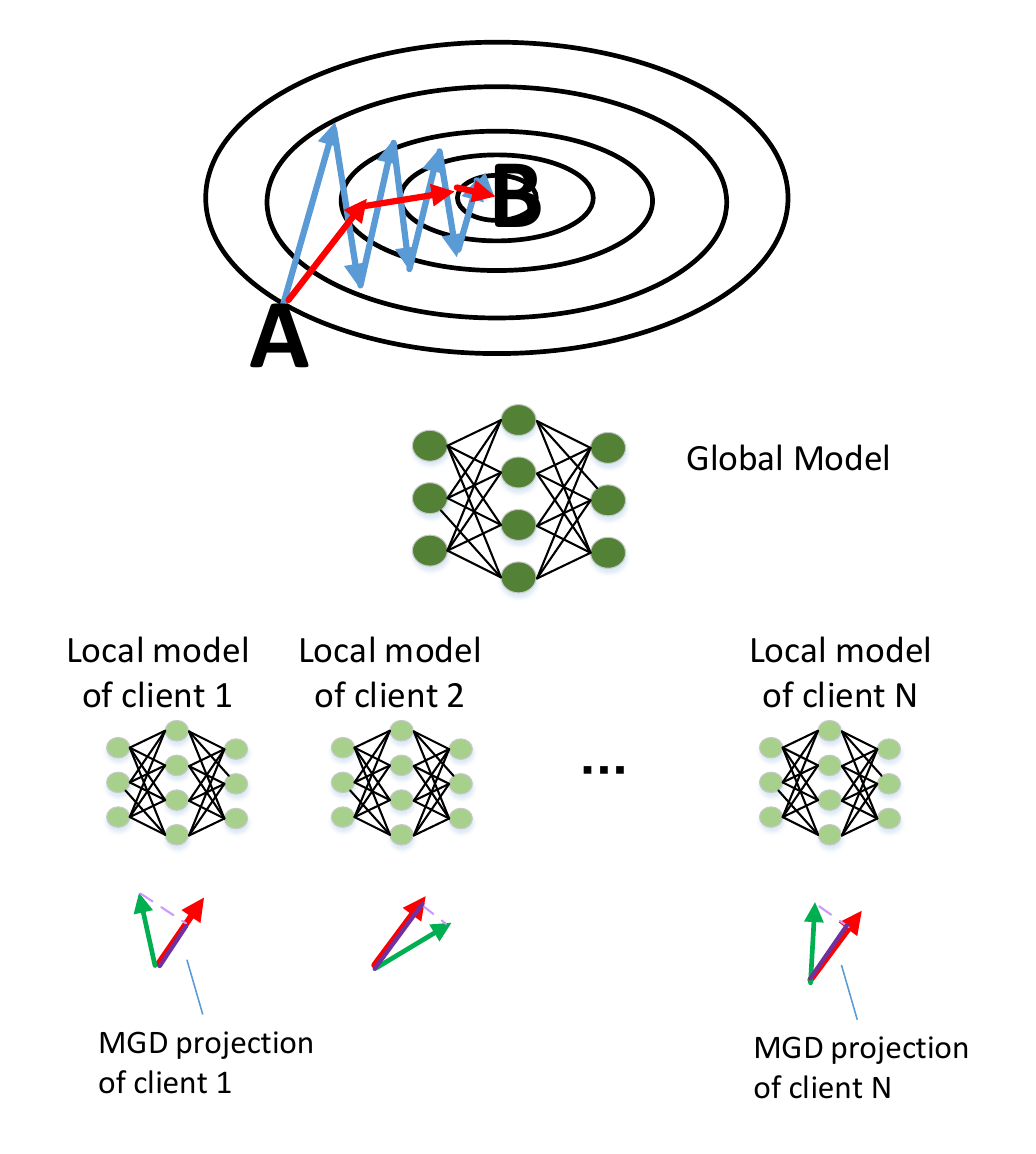}
 \caption{The designed Gradient Projection}
 \vspace{-3mm}
 \label{fig:MGP}
  \end{minipage}
  \label{fig:MGD}
\end{figure*}

In client selection, our goal is to choose $K$ clients that have made the most significant contribution to FL. To achieve this, we have developed an indicator called gradient projection (GP). GP is based on the assumption that the closer a client's locally calculated gradient is to the direction pointing towards the minimum loss, the greater its contribution to FL. We have selected Gradient Projection as our primary metric for data quality assessment over other similarity measures, based on the following reasoning: \textbf{First}, FL aims to optimize via gradient descent, prioritizing directional similarity over distance measures. Since determining the exact distance to the loss minimum is impractical, angle-based similarity is more relevant. \textbf{Second}, adhering to FL's privacy and efficiency, Cosine Similarity can be a candidate due to its focus on vector angles. We opted for Gradient Projection over Cosine Similarity due to its comprehensive consideration of both angular direction and magnitude. Alignment of gradient descent directions suggests data relevance but training only on similar data isn't always best. Introducing diversity helps escape local minima and achieve a global optimum. Using gradient magnitude in client selection boosts diversity, implicitly weighting client contributions, and valuing clients with significant differences from previous data, which can improve outcomes.

In order to simulate the direction from the starting point to the optimal point of the loss function, we utilize the global momentum-based gradient. Momentum-based gradient descent (MGD) is a method that improves the convergence rate of gradient descent (GD) by incorporating a momentum term into the gradient updating process. Momentum-based gradient descent maintains the impact of prior update directions when updating model parameters, which helps to smooth out the noise in stochastic gradients. Then, we can determine the direction that leads towards the minimum loss. Specifically, let $w^t$ denote the global model parameter at round t. For GD, the update reduction of the parameter is $\eta \nabla F(w^{t-1})$, which is only based on the gradient of $w^{t-1}$. This can result in an oscillating update path, as the update direction of GD always follows gradient descent.

Conversely, MGD's parameter update reduction is a combination of $\eta \nabla F(w^{t-1})$ and $\gamma(w^{t-2}-w^{t-1})$. The model parameter of the i-th client at round $t$ is computed as follows: 

\begin{equation}
\label{eq:MGD1}
d_i^t = \gamma d_i^{t-1} + \nabla F(w_i^{t-1})
\end{equation}
\begin{equation}
\label{eq:MGD2}
w_i^t = w_i^{t-1} - \eta d_i^t    
\end{equation}
where $d_i^t$ is the momentum term of the i-th client, $0 < \gamma <1$ is the momentum attenuation factor, $\eta > 0$ is the learning step size.

In Fig. \ref{fig:MGDvsGD}, we can observe that GD exhibits an oscillating update path and requires more iterations to reach the optimal point. In contrast, MGD, utilizing the momentum term, effectively deviates the parameter update direction towards the optimal decline and mitigates the oscillation caused by GD.


In the round $t$, we evaluate the gradients from each client and compare them to those computed by MGD and FedAvg for the global model in the $(t-1)$-th round. We hypothesize that if a client's gradient closely aligns with the global model's gradient, its contribution to model training is more substantial. Hence, we employ the projection of the gradient, which is calculated based on the local data of client $i$, as a metric to evaluate the contribution of client $i$ to FL training. Let $w_i^t$ denotes the local model parameter of client $i$ at round $t$, the gradient projection of client $i$ is computed by:

\begin{equation}
\label{eq:MGP}
\begin{aligned}
   c_i^t = \frac{\nabla F(w_i^t) \cdot \nabla F(w^{t-1})}{|\nabla F(w^{t-1})|}  
\end{aligned}
\end{equation}

\subsection{Gradient Projection Confidence Bound}

While GP can be used to determine the best clients for selection, solely focusing on this metric has limitations. Overlooking clients whose gradients do not align perfectly with the global model but still possess valuable information can hinder the model's ability to generalize across diverse data distributions. A bias towards clients with similar data distributions and close alignment may limit diversity and hinder overall performance improvement. Additionally, not considering clients with highly aligned gradients disregards opportunities for faster convergence and valuable insights.
          
When considering the individual contributions of each client, the objective of client selection is to choose the $K$ clients from $\mathcal{N}$ who have made the greatest impact. However, when taking into account the relationships between clients, the goal of client selection shifts to selecting $K$ clients from all possible permutations (a total of $C_{|\mathcal{N}|}^K$ options), and identifying the set of clients that contributes the most to FL training. To address the challenge, we incorporate an Exploration and Exploitation (EE) mechanism called Gradient Projection Confidence Bound-based client selection. This mechanism promotes exploration, diversity, and exploitation by striking a balance between exploring diverse data distributions and exploiting alignment. We approach the task of selecting $K$ clients from a pool of $|\mathcal{N}|$ clients by treating it as analogous to choosing $K$ arms from a MAB scenario with $|\mathcal{N}|$ arms. 



In contrast to solely prioritizing the $K$ arms with the highest GP values, which already encapsulates collaborative training information, our selection process assigns probabilities to arms that have been chosen less frequently in previous iterations. By adjusting expectations and continuously experimenting, our aim is to discover the optimal combination of clients that maximizes the overall performance.

Specifically, we begin by creating a regret function as follows:
\begin{equation}
R(T_{\text{trial}}) \overset{\text{def}}{=}\mathbb{E}\left[\sum_{t=1}^{T_{\text{trial}}}\sum_{i=1}^{N}(\mu_{i^*}^t - \mu_i^t)\right]
\end{equation}
where $T_{trial}$ is the total number of trials, $i^*$ is the best selection at round $t$, $i$ is the selected client at round $t$, and $\mu_{i}^t$ is the reward of selecting $i$.

We first define $\mu_{i}^t$ as follows:
\begin{equation}
\tilde{c}_i^t = \frac{\exp(c_i^t)}{\sum_{i^\prime=1}^{N} \exp(c_{i^\prime}^t)}, \qquad \mu_{i}^t = \tilde{c}_i^t 
\end{equation}
where $\tilde{c}_i^t$ is a normalized GP value at round $t$. 

Inspired by Upper Bound Confidence, we defined the Gradient Projection Confidence Bound (GPCB) of arm $i$ as follows:
\begin{equation}
\label{eq:UCB_}
u_{i} =\tilde{\mu}_{i}^t + \alpha \times \sqrt{\frac{2 \ln{n}}{n_i}}
\end{equation}
where $\tilde{\mu}_{i}^t = (\sum_{\phi=1}^t \mu_\phi) / t$ and the parameter $\alpha$ serves as an exploration parameter. During the initial stages, it is common for the model's loss to oscillate, potentially resulting in a decrease in accuracy.  increasing the value of $\alpha$. To be more specific, the calculation formula for determining $\alpha$ is:
 \begin{equation}
 \label{eq:alpha}
 \alpha = \rho \times \frac{t}{T}
 \end{equation}
 where $t$ represents the current round number, $T$ denotes the total number of rounds, and $\rho$ is a hyper-parameter. We set $\rho = 1$ in our experiment.

Then we select the top $K$ clients with the highest GPCB values as our client set. At this point, the selected clients may have high GP values (the first term in Equation \ref{eq:UCB_} is high) or they may have been selected less frequently in the past (the second term in Equation \ref{eq:UCB_} is high).




\subsection{Federated Learning with Gradient Projection Confidence Bound-Based Client Selection}
In rare instances, the momentum-based gradient of the global model may not precisely point to the optimal point in the loss function, resulting in the possibility of inaccuracies in calculating client reward $\mu_{i}^t$. To address this issue, we have devised a mechanism to fine-tune the reward value $\mu_{i}^t$. Specifically, if the model's accuracy improves in round $t$ compared to round $t-1$, it provides confirmation that the reward value is accurate. In such cases, we amplify the reward value to increase its influence. Conversely, if the accuracy in round $t$ decreases compared to round $t-1$, it indicates a potential inaccuracy in the reward value. It then becomes crucial to reduce the reward value if an undesired client has been selected.  For each selected client $i$, the formula for adjusting the reward value is as follows:
\vspace{-2mm}
\begin{equation}
\label{eq:MGP_adjustion}
\mu_{i}^t = \tilde{c}_{i}^t \times \left\{ \begin{array}{cc}
    2 \times \exp \left( A^t-A^{t-1} \right), &  A^t \neq A^{t-1} \\
    \exp \left(F(w^t) - F(w^{t-1})\right), & otherwise
\end{array} \right.
\end{equation}
where $F(w^t)$ is the loss of the global model at round $t$. 
The complete procedure of {\OURS} is outlined in Algorithm \ref{alg:mgpcs2}. As observed from Algorithm \ref{alg:mgpcs2}, during the initialization phase, all clients compute gradients. However, in the training round, only the selected clients are required to compute their local gradients. This approach eliminates the necessity of computing local gradients for every client in each round, giving it a notable advantage over other client selection algorithms. Consequently, our proposed {\OURS} method conserves client computing resources and substantially reduces the time taken for each round.



\begin{algorithm}[ht]
\caption{Federated Learning with Gradient Projection Confidence Bound-Based Client Selection}
\label{alg:mgpcs2}
\SetAlgoLined
\SetKwInput{Input}{Input}
\SetKwInput{Output}{Output}
\SetKwInput{Parameters}{Parameters}
\Input{The set of all clients $\mathcal{N}$ \\
\hspace{30pt} Initial parameters of the global model $w^0$}
\Output{The global model weight vector after $T$ rounds' training $w^T$}
\Parameters{$t \in T$ \\
\hspace{50pt} The learning step size of MGD $\eta$}
\text{//Initialization} \\
$C \gets [0] \times |\mathcal{N}|$\;
\ForEach{client $i \in U$ in parallel}{
   $w_i^0 \gets w^0$\;
   compute $w_i^0$ based on Equation \ref{eq:MGD2}\;
   compute $c_i^0$ based on Equation \ref{eq:MGP}\;
   $C_i \gets c_i^0$\;
}
$\mathcal{S}_1 \gets sort(C,K,des)$\;
$w_1 \gets FedAvg(w^0,\mathcal{S}_0,C)$\;
\text{//Training} \\
\For{$t = 1,...,T$}{
  \ForEach{client $i \in \mathcal{S}_t$ in parallel}{
     compute $w_i^t$ based on Equation \ref{eq:MGD2}\;
	 compute $c_i^t$ based on Equation \ref{eq:MGP}\;
     compute $\mu_i^t$ based on Equation \ref{eq:MGP_adjustion}\;
     compute $u_i^t$ based on Equation \ref{eq:UCB_}\;
	 $C_i \gets c_i^t$\;
  }
  $\mathcal{S}_t \gets sort (C,K,des)$\;
  $w^t \gets FedAvg(w^{t-1},\mathcal{S}_t,C)$\;
}
\textbf{return} $w^T$\;
\end{algorithm}

\subsection{Theoretical Analysis}
\label{sec:TA}

\begin{assumption}
Each agent $i$ has a quality $\mu_{i}$ drawn from a distribution with mean $\overline{\mu}_{i}$ and variance $\sigma_i^2$.
\end{assumption}

\begin{assumption}
The reward factor $\mu_{i}$ are bounded in the range $[0,1]$.
\end{assumption}

\begin{theorem}
\label{theorem:bound}
Under the above assumptions, the regret of the MAB algorithm in the IID setting, with the exploration phase, is bounded as follows:

\[
\mathbb{E}[R_t] \leq \frac{t}{1 - (t+1) \exp\left(-\frac{\tau}{2}\right)} \exp\left(-\frac{\tau}{2}\right)
\]

where $t$ is the number of rounds, $\tau = \frac{2 \ln n}{n_i}$ is the exploration term, with $n_i$ denoting the number of times arm $i$ has been selected until round $t$.
\end{theorem}
Due to space limitations, we present a proof outline for Theorem \ref{theorem:bound} as below. A detailed proof is available in the appendix section.

We consider the IID setting where the exploration phase is carried out. Each agent is explored for $\tau = \frac{2 \ln n}{n_i}$ rounds, where $n_i$ is the number of times arm $i$ has been selected until round $t$.
We analyze the expected regret of the MAB algorithm in this setting.
By setting the exploration parameter $\tau$ as mentioned above, we can show that after $\tau$ rounds, the average value of $(\hat{\mu}_i^{t})^+$ (the upper confidence bound estimate of agent $i$'s quality) and the average value of $\hat{\mu}_i^{t}$ (the estimate of agent $i$'s quality) for the agents in the selected super-arm $a_t$ are both less than $\epsilon_2$ with high probability.
We define $\mathcal{S}_t$ as the set of winner nodes at round $t$. Let $\beta$ be the maximum mean quality among all agents, i.e., $\beta = \mathop{max}\limits_i \mu_i$. If the number of agents $i$ in $\mathcal{S}_t$ with $(\hat{\mu}_i^{t})^+ \geq \beta + \epsilon_2$ is at least $s_t/2$, we can show that the probability of the average quality of the super-arm $a_t$ being less than $\beta - \epsilon_1$ is small.
We calculate the probability $p_t$ of selecting the correct super-arm at round $t$ and derive a lower bound for $p_t$ by considering the probabilities of different events.
By setting $\delta = (t+1) \exp\left(-\frac{\tau}{2}\right)$, we can show that $p_t \geq 1 - \delta$.
Using the derived lower bound for $p_t$, we obtain an upper bound for the expected regret $\mathbb{E}[R_t]$ of the MAB algorithm.
The regret bound is expressed as $\mathbb{E}[R_t] \leq \frac{t}{1 - (t+1) \exp\left(-\frac{\tau}{2}\right)} \exp\left(-\frac{\tau}{2}\right)$.

\section{Convergence Analysis}
\label{sec:TA}

We have the following assumptions:

\begin{assumption}
Each agent $i$ has a quality $\mu_i$ drawn from a distribution with mean $\overline{\mu}_i$ and variance $\sigma_i^2$.
\end{assumption}

\begin{assumption}
The reward factors $\mu_i$ are bounded in the range $[0,1]$.
\end{assumption}

We aim to prove the following theorem:

\begin{theorem}
Under the above assumptions, the regret of the MAB algorithm in the IID setting, with the exploration phase, is bounded as follows:
$$
\mathbb{E}[R_t] \leq \frac{t}{1 - (t+1) \exp\left(-\frac{\tau}{2}\right)} \exp\left(-\frac{\tau}{2}\right)
$$
where $t$ is the number of rounds, $\tau = \frac{2 \ln n}{n_i}$ is the exploration term, and $n_i$ denotes the number of times arm $i$ has been selected until round $t$.
\end{theorem}

\begin{proof}
In the exploration phase of the MAB algorithm, each agent $i$ is explored for $\tau = \frac{2 \ln n}{n_i}$ rounds. We want to analyze the expected regret $\mathbb{E}[R_t]$ of the algorithm in this setting.

First, we consider the exploration phase and show that after $\tau$ rounds, the average value of $(\hat{\mu}_i^{t})^+$ (the upper confidence bound estimate of agent $i$'s quality) and the average value of $\hat{\mu}_i^{t}$ (the estimate of agent $i$'s quality) for the agents in the selected super-arm $a_t$ are both less than $\epsilon_2$ with high probability. This can be expressed as:

\[
\mathbb{P}\left(\frac{1}{\tau}\sum_{t=1}^{\tau}(\hat{\mu}_i^{t})^+ \leq \epsilon_2\right) \geq 1 - \exp\left(-\frac{\tau}{2}\right)
\]

\[
\mathbb{P}\left(\frac{1}{\tau}\sum_{t=1}^{\tau}\hat{\mu}_i^{t} \leq \epsilon_2\right) \geq 1 - \exp\left(-\frac{\tau}{2}\right)
\]

Next, we define $\mathcal{S}_t$ as the set of winner nodes at round $t$. Let $\beta$ be the maximum mean quality among all agents, i.e., $\beta = \max_i \mu_i$. Let $s_t$ be the cardinality of the selected super-arm. If the number of agents $i$ in $\mathcal{S}_t$ with $(\hat{\mu}_i^{t})^+ \geq \beta + \epsilon_2$ is at least $s_t/2$, we can show that the probability of the average quality of the super-arm $a_t$ being less than $\beta - \epsilon_1$ is small. This can be expressed as:

\[
\mathbb{P}\left(\frac{1}{s_t}\sum_{i \in \mathcal{S}_t}(\hat{\mu}_i^{t})^+ \geq \beta + \epsilon_2\right) \geq \frac{s_t}{2}
\]

\[
\mathbb{P}\left(\frac{1}{s_t}\sum_{i \in \mathcal{S}_t}\hat{\mu}_i^{t} \leq \beta - \epsilon_1\right) \leq \exp\left(-\frac{s_t}{8}\right)
\]

Now, let's calculate the probability $p_t$ of selecting the correct super-arm at round $t$ and derive a lower bound for $p_t$ by considering the probabilities of different events.

The event that the true best super-arm is not selected can be expressed as the event that either the average quality of the selected super-arm is less than $\beta - \epsilon_1$ or the average quality of a non-selected super-arm is greater than $\beta + \epsilon_2$. This can be written as:

\[
\mathbb{P}\left(\frac{1}{s_t}\sum_{i \in \mathcal{S}_t}\hat{\mu}_i^{t} \leq \beta - \epsilon_1 \text{ or } \exists i \notin \mathcal{S}_t : \frac{1}{s_t}\sum_{i \in \mathcal{S}_t}(\hat{\mu}_i^{t})^+ \leq \beta + \epsilon_2\right)
\]

Using the union bound, we can bound the probability of not selecting the correct super-arm as:

\[
\mathbb{P}\left(\frac{1}{s_t}\sum_{i \in \mathcal{S}_t}\hat{\mu}_i^{t} \leq \beta - \epsilon_1 \text{ or } \exists i \notin \mathcal{S}_t : \frac{1}{s_t}\sum_{i \in \mathcal{S}_t}(\hat{\mu}_i^{t})^+ \leq \beta + \epsilon_2\right)\]
\[\leq \exp\left(-\frac{s_t}{8}\right) + \exp\left(-\frac{\tau}{2}\right)
\]

Let $\delta = \exp\left(-\frac{s_t}{8}\right) + \exp\left(-\frac{\tau}{2}\right)$. Then, the probability $p_t$ of selecting the correct super-arm at round $t$ is given by $p_t = 1 - \delta$.

Using the derived lower bound for $p_t$, we can obtain an upper bound for the expected regret $\mathbb{E}[R_t]$ of the MAB algorithm. Therefore, the expected regret can be bounded as:

\[
\mathbb{E}[R_t] \leq \frac{1}{p_t}(\beta + \epsilon_1) \leq \frac{1}{1 - \delta}(\beta + \epsilon_1)
\]

Finally, we substitute the value of $\delta$ and simplify the expression to obtain the desired regret bound:

\[
\mathbb{E}[R_t] \leq \frac{t}{1 - (t+1) \exp\left(-\frac{\tau}{2}\right)} \exp\left(-\frac{\tau}{2}\right)
\]

This completes the proof.
\end{proof}

\section{Experiment Settings}
\subsection{Dataset and Data Distribution}
We conducted federated learning experiments on two benchmark datasets: FEMNIST and CIFAR-10.
Federated Extended MNIST (FEMNIST) is a dataset specifically crafted for federated learning research, offering a realistic and challenging environment for developing and testing machine learning models. It is based on the Extended MNIST dataset, which includes both handwritten digits (0-9) and letters (A-Z, a-z). The dataset is part of the LEAF benchmark suite, which includes a range of datasets and evaluation tools to facilitate reproducible research and comparison of federated learning algorithms.
CIFAR-10 is a high-quality dataset for image classification tasks, consisting of 60,000 32x32 color images categorized into 10 classes, each containing 6,000 images. The original dataset is evenly split into 50,000 training images and 10,000 test images, and it includes a diverse set of objects such as airplanes, cars, birds, cats, deer, dogs, frogs, horses, ships, and trucks. 


The partitioning of the two datasets is designed to emulate a federated learning environment, wherein each device or user contributes data originating from a distinct source. The summary statistics for these datasets are presented in Table \ref{tb:dataset_appendix}, which encompasses details such as the count of devices, the total number of samples, the average number of samples per device, and the associated standard deviation.


\begin{table}[ht]
\centering
\begin{tabular}{c|c|c|c}
\hline
\hline
\multicolumn{2}{c|}{}  &\textbf{FEMINST}& \textbf{CIFAR-10}\\
\hline
\multicolumn{2}{c|}{\textbf{Client Count}} & 3,550 & 100 \\
\hline
\multicolumn{2}{c|}{\textbf{Sample Count}} & 805,263 & 60,000 \\
\hline
\multirow{2}*{\textbf{Sample per Client}} & Mean & 226.83 & 946.8 \\
\cline{2-4}
        & Stdev & 88.94 & 256.04 \\
\hline
\hline
\end{tabular}
\caption{Statistics of dataset}
\label{tb:dataset_appendix}
\end{table}



We investigate three different heterogeneous settings to analyze the effects of data partitioning and client heterogeneity on the performance of federated learning:

\begin{enumerate}
\item \textbf{2 shards per client (2SPC):} 
The data is sorted by labels and divided into 200 shards, where each shard contains samples from the same label. The shards are randomly allocated to clients, with each client receiving two shards. As all shards have the same size, the data partition is balanced, ensuring that all clients have the same dataset size. We select K = 5 clients in each round within this setting.

\item \textbf{1 shard per client (1SPC):} This setting is similar to the 2SPC setting, but each client is assigned only one shard, representing the highest level of data heterogeneity. Each client contains data from a single label, resulting in a balanced partition. In each round within this setting, we select K = 10 clients.

\item \textbf{Dirichlet Distribution with $\zeta = 0.2$ (Dir):} 
For each client, denoted as $i$, we independently sample the data distribution $q_i$ from a Dirichlet distribution, which can be formulated as $q_i \sim \text{Dir}(\zeta p)$. Here, $p$ represents the prior label distribution, and $\zeta \in \mathbb{R}^+$ is the concentration parameter of the Dirichlet distribution. By combining the $q_i$ values for all clients, we obtain a fraction matrix $Q = [q_1, q_2, ..., q_n]$. The dataset size on each client is denoted as $x = [x_1, x_2, ..., x_N]^T$, and we determine it through a quadratic programming solution. We minimize $\lVert x \rVert_2$ while ensuring that the matrix product $Qx$ equals the desired distribution $d$, where $d$ represents the number of data samples for each label. This minimization helps avoid cases where the data distribution becomes overly concentrated on a small fraction of clients, as it would make the client selection problem trivial. We set the concentration parameter $\zeta$ to 0.2 to create the Dirichlet Distribution with $\zeta = 0.2$ (Dir) setting, resulting in an unbalanced data partition. This setting introduces varying proportions of different labels across clients, generating heterogeneity in the dataset partition. In each round of this setting, we select five clients ($K = 5$) to participate in the federated learning process.
\end{enumerate}

\subsection{Model Construction and Hyperparameters}

\subsubsection{Model Construction and Hyperparameters in FEMNIST}
For the FEMNIST dataset, we adopt an MLP (Multi-Layer Perceptron) architecture with two hidden layers, consisting of 64 and 30  units, respectively.
In all three heterogeneous settings, the local batch size is set to 64, and the number of local iterations is set to 20. We initialize the learning rate ($\eta$) as 0.005. We utilize an SGD (Stochastic Gradient Descent) optimizer with a weight decay of 0.0001 and momentum of 0.1. The data is divided among $N = 100$ clients, with a participation fraction ($K$) of 10 for the 1SPC setting and 5 for the 2SPC and Dir settings.

\subsubsection{Model Construction and Hyperparameters in CIFAR-10}
For the CIFAR-10 dataset, we employ a CNN (Convolutional Neural Network) architecture with three convolutional layers, consisting of 32, 64, and 64 kernels, respectively, followed by a fully-connected layer with 64 units. In all three heterogeneous settings, the local batch size is set to 50, and the number of local iterations is set to 40. We use a fixed learning rate ($\eta$) of 0.01 without any learning rate decay and a weight decay of 0.0003 for the SGD optimizer. The total number of clients and the client participation fraction are consistent with those used in the FEMNIST dataset.

By leveraging these specific hyperparameters and employing the respective neural models, we conducted federated learning experiments on the FEMNIST and CIFAR-10 datasets under different heterogeneous settings.


%% file: 4Evaluation.tex
\section{Experiment}
In this section, we evaluate the performance of {\OURS}. To achieve this, we create three non-independent and identically distributed data environments to compare our proposed algorithm's performance with various baselines. We assess the FL test accuracy under various client selection methods and evaluate the time required for each approach. Furthermore, we conduct an ablation study to gauge the effectiveness of the exploration-exploitation components incorporated in {\OURS}.

\subsection{Evaluation Setup}


1) \textit{Datasets}: We have conducted experiments on two real-world datasets, namely CIFAR-10 \cite{krizhevsky2009learning} and Federated-MNIST (FEMNIST) \cite{caldas2018leaf}. Following \cite{tang2022fedcor}, we used three different heterogeneous data partitions: 2 shards per client (2SPC), 1 shard per client (1SPC), and Dirichlet Distribution (Dir). In the 2SPC setting, the data is sorted based on their labels and divided into 200 shards such that all data within a single shard share the same label. Similarly, the 1SPC setting is identical to the 2SPC setting, except that each client only has one shard, meaning they possess data from only one label. Finally, the Dirichlet distribution is an unbalanced data partitioning method. For more details on our experimental setup, please refer to the Appendix. 

2) \textit{Models and Parameters}: In the FEMINST experiment, the MINST dataset is divided among 3550 clients, with an average of 226 samples per client. In contrast, in the CIFAR-10 experiment, the data is distributed across 100 clients, with an average of 947 samples per client. We have set the number of selected clients $K=10$ for 1SPC and $K=5$ for 2SPC and Dir. For classification purposes, we use a CNN network on both datasets. More detailed information about the datasets and model construction can be found in the Appendix.




3) \textit{Baselines}: To validate our proposed GPFL, The baselines we employed for comparison encompassed the random method, as well as the latest and representative research findings, namely FedCor\cite{tang2022fedcor} and Pow-d \cite{cho2022towards}. FedCor first models the loss correlations between clients using a Gaussian Process, then derives a set of $K$ clients that significantly reduce the expected global loss. Pow-d demonstrates that prioritizing clients with higher local loss leads to quicker convergence of errors. The random method randomly selects $K$ clients to participate in FL training. In particular, for the client selection process, we utilized the open-source code from https://github.com/Yoruko-Tang/FedCor to obtain the client results of FedCor, Pow-d, and Random. Subsequently, we aggregated the updated models from the selected clients using FedAvg \cite{DBLP:conf/iclr/LiHYWZ20}.

\subsection{Result Comparison}


\begin{table*}[ht]
\begin{tabular}{c|c|ccc|ccc}
\hline
\hline
\multirow{2}{*}{Iteration} & \multirow{2}{*}{Method} & \multicolumn{3}{c|}{FEMINST}                                                  & \multicolumn{3}{c}{CIFAR-10}                                      \\ \cline{3-8} 
                           &                         & 1SPC                 & 2SPC                            & Dir                  & 1SPC                & 2SPC                 & Dir                  \\ \hline
\multirow{4}{*}{15\%}      & Random                  & 0.4895               & 0.5378                          & 0.4859               & 0.1347              & 0.3493               & 0.2849               \\
                           & Pow-d                   & 0.4826               & 0.5824                          & 0.5802               & 0.1515              & 0.2867               & \textbf{0.3089}      \\
                           & FedCor                  & 0.6646               & \textbf{0.6971}                         & 0.5635               & \textbf{0.1526}     & 0.2455               & 0.2455               \\
                           & GPFL(ours)              & \textbf{0.6789}      & 0.6879               & \textbf{0.6268}      & 0.1000            & \textbf{0.3659}      & 0.2941               \\ \hline
\multirow{4}{*}{50\%}      & Random                  & 0.4331               & 0.5367                          & 0.5575               & 0.2219              & 0.3008               & 0.3655               \\
                           & Pow-d                   & 0.4865               & 0.6327                          & 0.4903               & \textbf{0.2885}     & 0.389                & 0.3419               \\
                           & FedCor                  & 0.6717               & 0.6907                          & 0.6814               & 0.2047              & 0.4081               & 0.4081               \\
                           & GPFL(ours)              & \textbf{0.6956}      & \textbf{0.7741}                 & \textbf{0.7411}      & 0.2352              & \textbf{0.5202}      & \textbf{0.4183}      \\ \hline
\multirow{4}{*}{100\%}     & Random                  & 0.5020 $\pm$ 0.08          & 0.6001 $\pm$ 0.11                     & 0.5451 $\pm$ 0.08          & 0.2667 $\pm$ 0.08         & 0.5257 $\pm$ 0.06          & 0.5179 $\pm$ 0.05          \\
                           & Pow-d                   & 0.4801 $\pm$ 0.15          & \multicolumn{1}{l}{0.5859 $\pm$ 0.08} & 0.5831 $\pm$ 0.16          & \textbf{0.3110 $\pm$ 0.08} & 0.5613 $\pm$ 0.18          & 0.4989 $\pm$ 0.18          \\
                           & FedCor                  & 0.5714 $\pm$ 0.12          & \multicolumn{1}{l}{0.6846 $\pm$ 0.04} & 0.6359 $\pm$ 0.09          & 0.2719 $\pm$ 0.10         & 0.4759 $\pm$ 0.10          &0.4821 $\pm$ 0.10          \\
                           & GPFL(ours)              & \textbf{0.7703 $\pm$ 0.04} & \textbf{0.7780 $\pm$ 0.04}            & \textbf{0.7370 $\pm$ 0.05} & 0.2822 $\pm$ 0.10         & \textbf{0.5755 $\pm$ 0.05} & \textbf{0.5346 $\pm$ 0.06} \\ \hline
                           \hline
\end{tabular}
\caption{Test Accuracy Comparison. The FEMINST and CIFAR-10 datasets were trained for a total of 500 and 2000 rounds, respectively. The final results reported at 100\% represent the average test accuracy achieved during the last 10 rounds, while the values following '$\pm$' indicate the maximum deviation between the test accuracy of the last 10 rounds and their respective average.}
\label{table:overall_acc}
\end{table*}

\begin{figure*}[!t]
\centering
\subfloat[1SPC]{    
\includegraphics[width=0.3\textwidth, trim=0 0 0 0, clip]{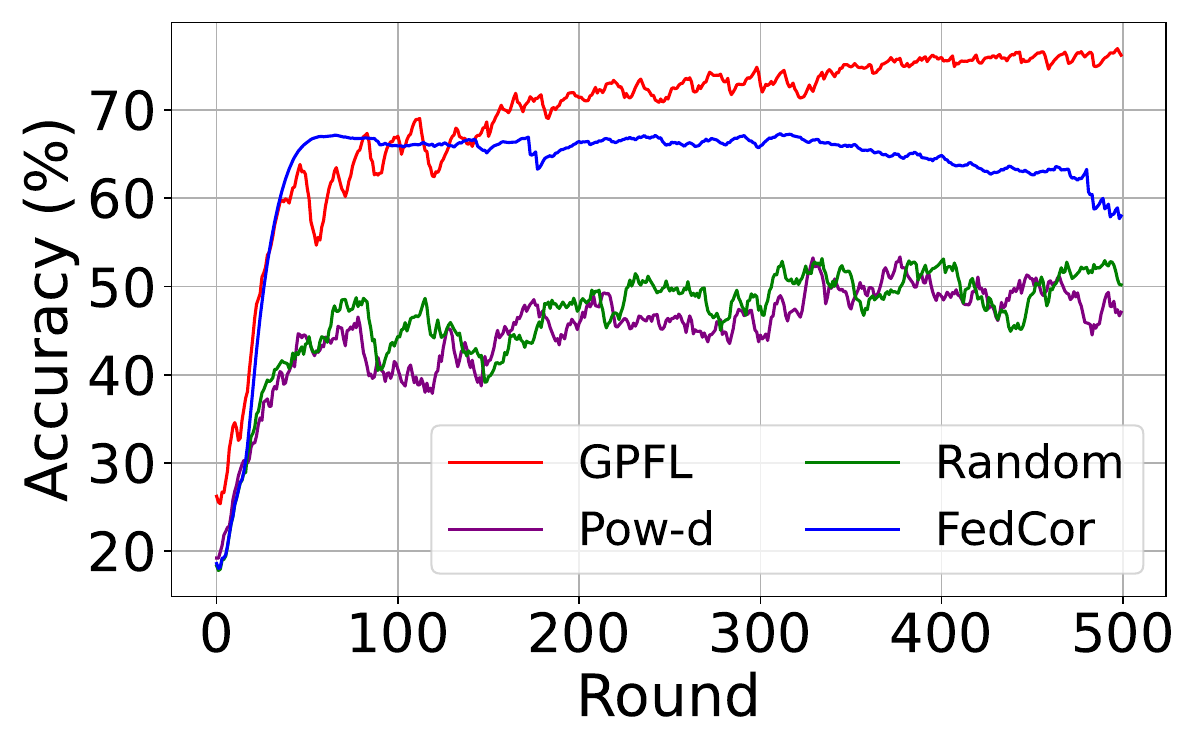}
\label{fmnist_1shard}
}
\subfloat[2SPC]{     
\includegraphics[width=0.3\textwidth, trim=0 0 0 0, clip]{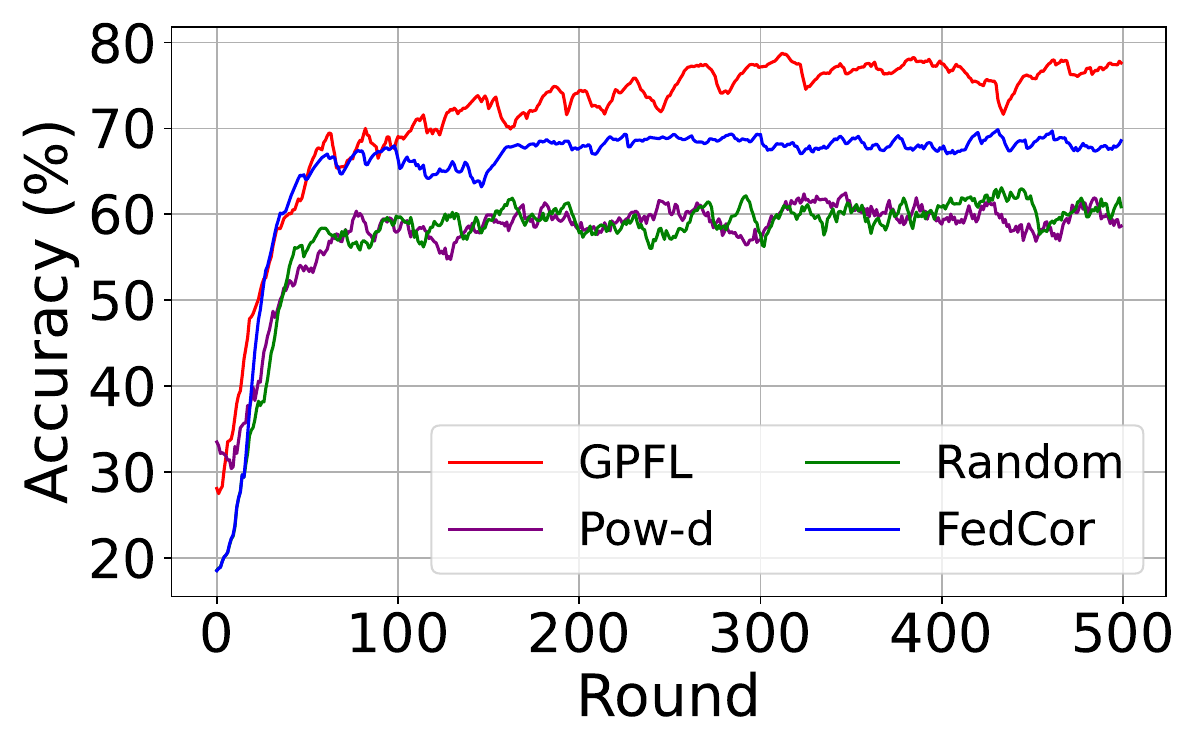}
\label{overlap2}
}
\subfloat[Dir]{     
\includegraphics[width=0.3\textwidth, trim=0 0 0 0, clip]{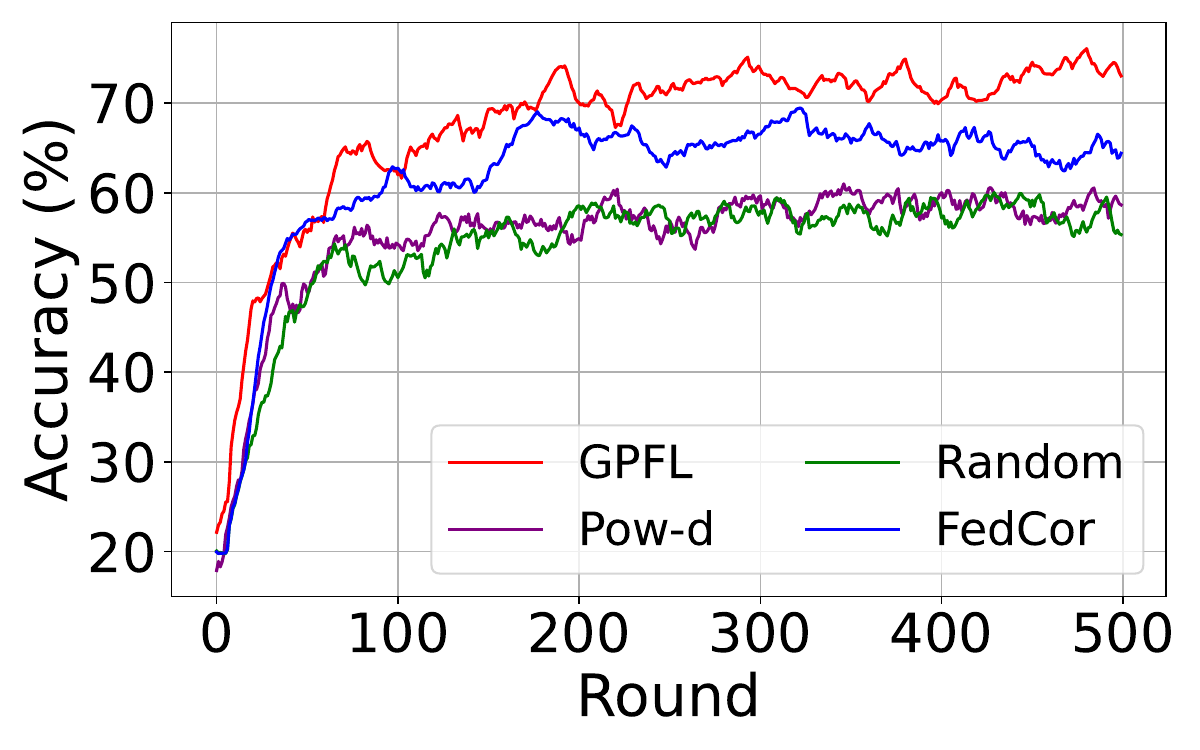}
\label{fmnist_dir}
}
\caption{Performance Comparison on FEMINIST}
\vspace{-3mm}
\label{Fig:fmnist}
\end{figure*}


In this section, we compare the performance of {\OURS} with existing client selection methods. To ensure a fair comparison, all of the methods utilize FedAvg for model aggregation, resulting in the primary difference in their experimental outcomes being the client selection phase. The results on FEMINIST and CIFAR-10 datasets are presented in Table \ref{table:overall_acc}, while Figure \ref{Fig:fmnist} illustrates the accuracy fluctuations of GPFL and other baseline algorithms across three distinct settings over multiple rounds.

According to Table \ref{table:overall_acc} and Figure \ref{Fig:fmnist}, we have the following observations: 1) {\OURS} consistently surpasses other methods on the FEMINST dataset. Across various data distributions, FEMINST quickly achieves high test accuracy, and in the end, our algorithm outperforms the highest baseline by 19.89\%, 9.34\%, and 10.11\% on 1SPC, 2SPC, and Dir distributions respectively. Additionally, {\OURS} exhibits the smallest fluctuation in test accuracy over the last 10 rounds, with a difference of no more than 4\% between the maximum and minimum values and the mean. This demonstrates that the proposed gradient projection (GP) can accurately measure data quality in Non-IID data, and when combined with our EE mechanism, it selects an appropriate client set. 2) On the CIFAR-10 dataset, {\OURS} still outperforms baseline methods under 2SPC and Dir settings, but it falls short of Pow-d under the 1SPC setting. We believe this is due to the fact that classifying animals in colored images on CIFAR-10 is a more challenging task than the classification task on the MINST dataset. The highly imbalanced 1SPC data distribution makes it difficult for FedAvg to achieve good results, which is also reflected in the large fluctuations observed in all algorithms during the last 10 rounds of this setting. While Pow-d, as a post-selection method, can select based on the running results of more clients and obtain slightly better experimental results, it comes at the cost of consuming more client computing resources. 3) FedCor has demonstrated impressive performance in the early rounds of FEMINST, but its performance on CIFAR-10 is even worse than random selection. We hypothesize that this is due to FedCor's strategy of avoiding selecting two clients with similar data. In simple tasks with sufficient data, this approach can yield good results. However, when the task becomes more complex and there is not enough data, combining this method with FedAvg makes it difficult to achieve good experimental results.


\subsection{Result Analysis}

\begin{figure}[ht]
\centering
\includegraphics[width=0.7\textwidth, trim=110 70 70 60, clip]{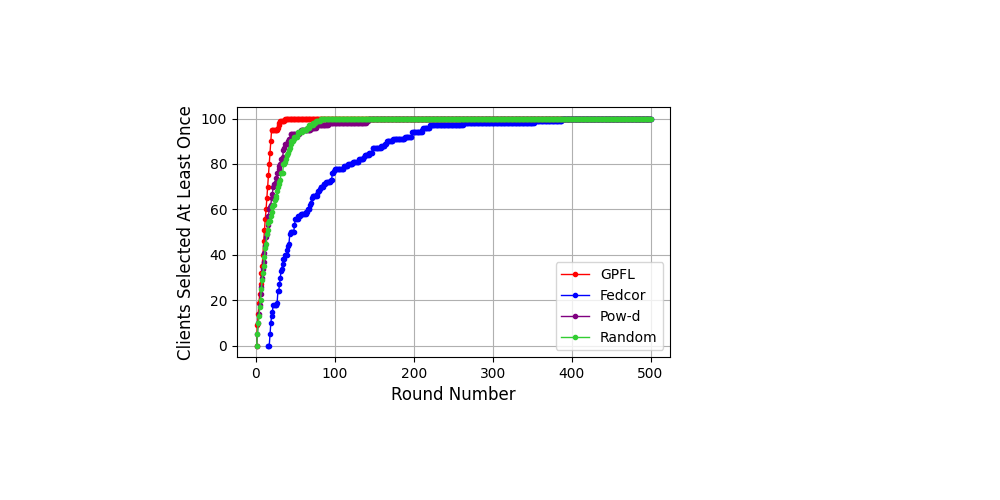}
\caption{Number of Clients Selected At Least Once Over Rounds}
\label{fig:atleastonce}
\end{figure}

\begin{figure*}[ht]
\centering
\subfloat[GPFL]{
\includegraphics[width=0.45\textwidth, trim=0 0 50 0, clip]{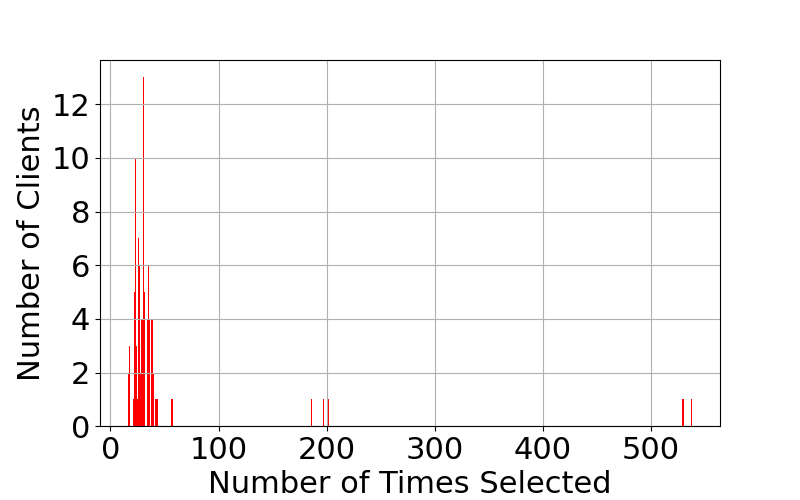}
\label{gpfl_fre}
}
\hfil
\subfloat[FedCor]{
\includegraphics[width=0.45\textwidth, trim=0 0 50 0, clip]{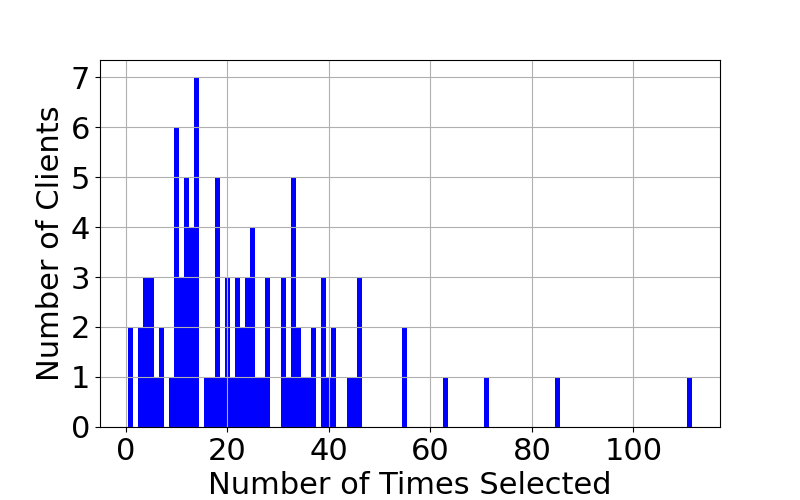}
\label{fedcor_fre}
}
\vfil
\subfloat[Pow-d]{
\includegraphics[width=0.45\textwidth, trim=0 0 50 0, clip]{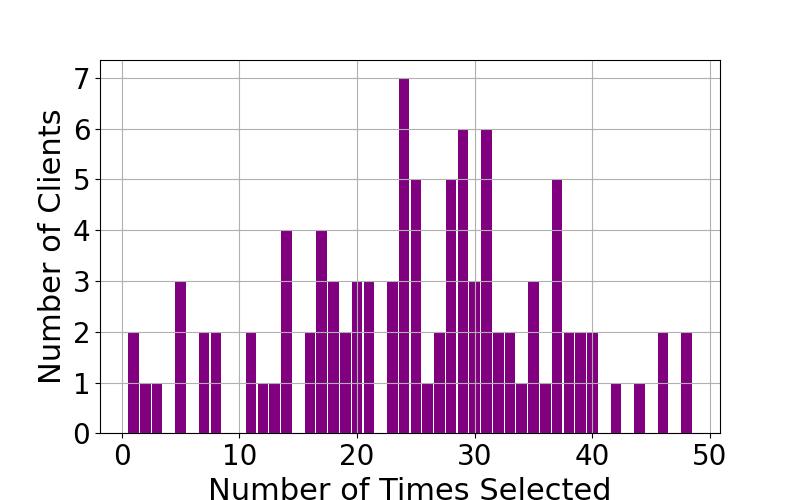}
\label{Pow-d_fre}
}
\hfil
\subfloat[Random]{
\includegraphics[width=0.45\textwidth, trim=0 0 50 0, clip]{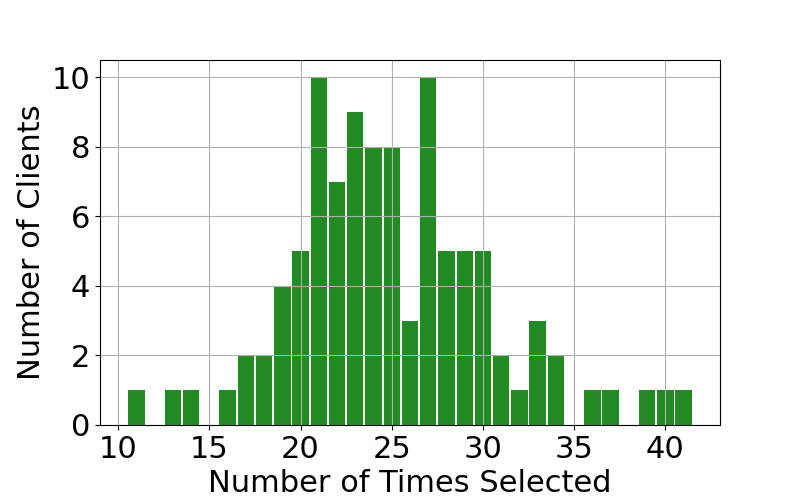}
\label{Random_fre}
}
\caption{Histogram for selection frequency}
\vspace{-3mm}
\label{Fig:histogram}
\end{figure*}

In this section, we delve deeper into the analysis of our experimental results, focusing on the characteristics of client selection exhibited by various algorithms. We conducted a pair of experiments designed to evaluate these characteristics. The first experiment monitored the proportion of clients selected at least once as the number of training rounds increased. It is imperative in Federated Learning to ensure that client selection algorithms do not consistently omit certain clients, as each client should ideally be chosen early on to gauge their unique contribution effectively. This particular experiment tracks how many rounds it takes for every algorithm to include all clients in the selection process at least once. The second experiment plotted the frequency of client selections against the total number of clients, providing insight into the propensity of the algorithms towards certain clients over others.

Illustrated in Figure \ref{fig:atleastonce}, we observe the progression of the percentage of clients selected across different training rounds. The x-axis denotes the training rounds, while the y-axis denotes the proportion of clients selected at least once per round. Our proposed GPFL algorithm impressively ensured that all clients were selected at least once within a mere 50 rounds. This performance significantly outpaced the nearly 100 rounds required by random and pow-d algorithms, with FedCor taking approximately 200 rounds to achieve equivalent coverage. The accelerated rate at which GPFL selects clients can be attributed to its strategic exploration and exploitation approach. In the early stages of training, when the element of exploitation is not yet stable, GPFL assigns a higher priority to exploration, favoring clients that have not been previously selected. This tactic enables a more rapid and comprehensive selection compared to random and pow-d methods. Such prioritization is particularly beneficial in federated learning scenarios where data heterogeneity is crucial, as it allows the model to leverage the full spectrum of available data early on, potentially reducing bias and enhancing overall performance. The prolonged selection process exhibited by FedCor—surpassing 200 rounds before including all clients—hints at a subset of non-selected clients that were overlooked during the initial rounds over an extended period.

The histogram in Figure \ref{Fig:histogram} presents the results of the second experiment, illustrating the frequency of client selection versus the number of clients. Here, the x-axis represents the frequency of selections, and the y-axis denotes the quantity of clients. Across all tested baselines, a roughly normal distribution is evident with an average around 20. However, GPFL diverges with a distinctive long tail reaching up to around 500, suggesting that a small group of clients were selected considerably more frequently. This pattern indicates that GPFL strikes a balance between selecting a diverse range of clients and repeating selections for a chosen few. Conversely, 'Fedcor' displays a long tail extending to about 120 selections, indicating a tendency to repeatedly choose a limited subset of clients. Pow-d and Random exhibit more conventional normal distributions, with Pow-d showing a flatter curve centered on (24, 7) and Random peaking around (25, 10). The distinctive feature of GPFL lies in its ability to select a broad range of clients while also favoring repetition for certain clients. This balance ensures a diverse representation in training while allowing for deeper exploration of influential client data. This strategy potentially contributes to the effectiveness of GPFL in leveraging the full spectrum of available client data for model training, leading to enhanced model robustness and generalization capability.

\subsection{Time Comparison}
\begin{figure}[htbp]
    \centering
\subfloat[FEMNIST]{
\includegraphics[width=0.48\textwidth, trim=0 0 0 0, clip]{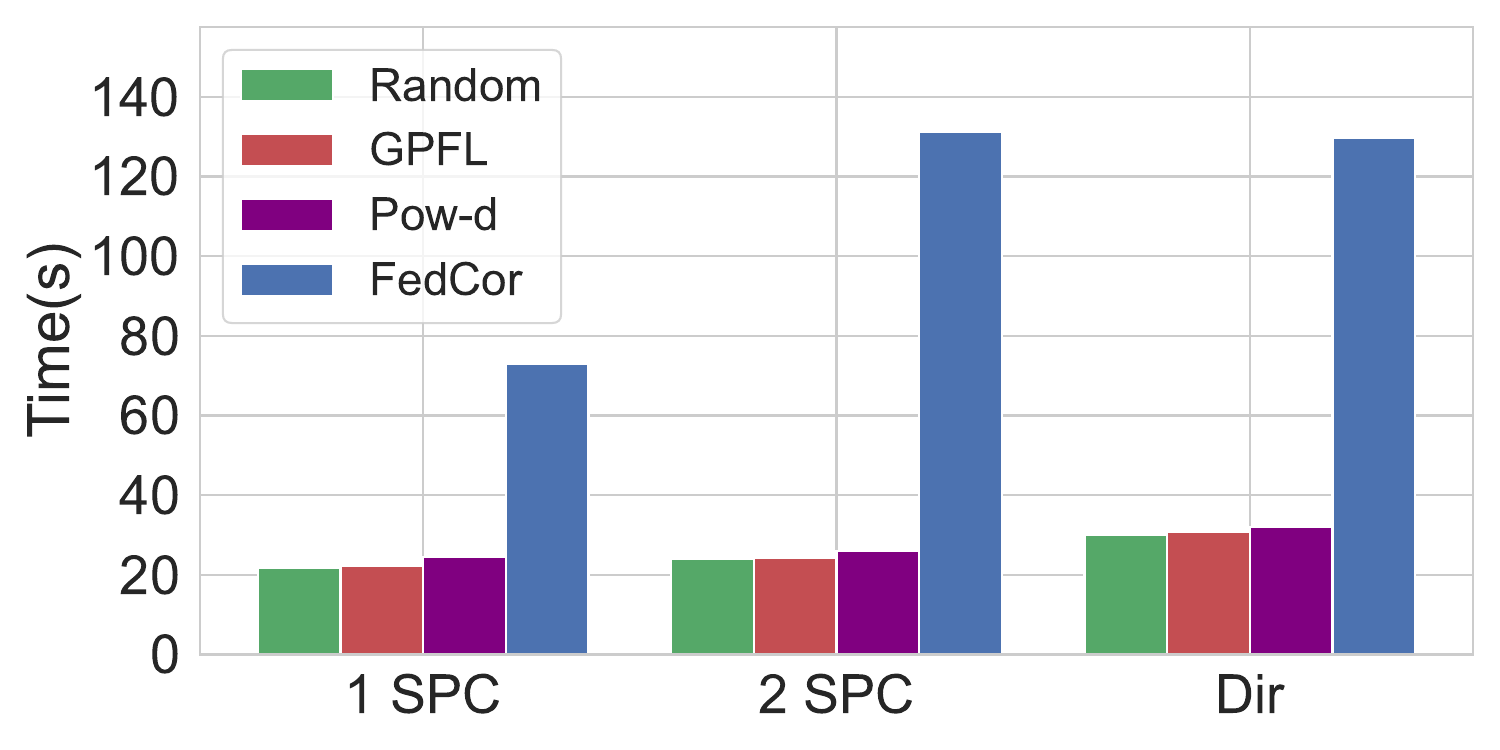}
\label{fig:sub1}
}
\hfill
\subfloat[CIFAR-10]{
\includegraphics[width=0.48\textwidth, trim=0 0 0 0, clip]{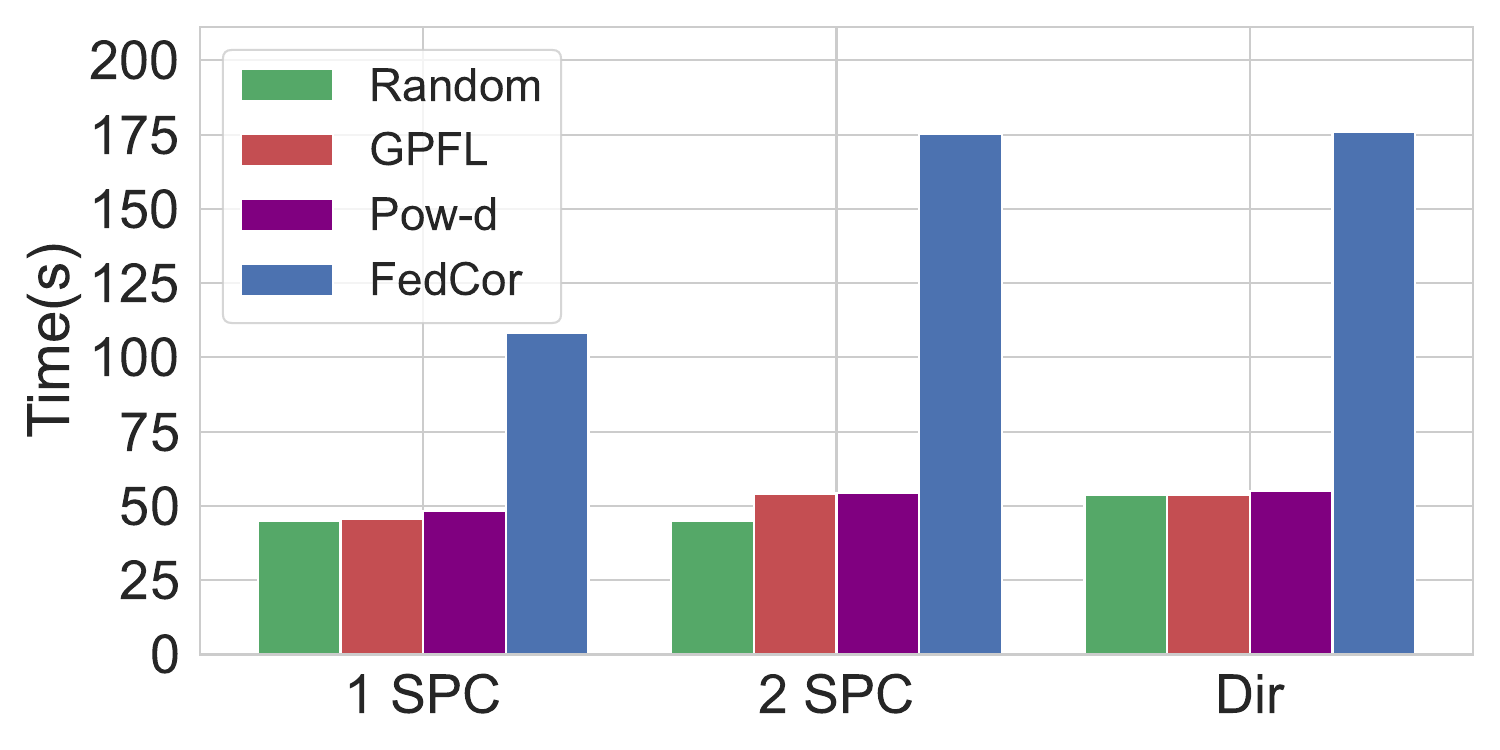}
\label{fig:sub2}
}
\caption{Training Time Comparison}
\label{fig:time}
\end{figure}

In this section, we compare the training time required by {\OURS} and other baseline methods. Specifically, we measure the time taken to run each method for 200 rounds and present the results in Figure \ref{fig:time}. As can be seen, our proposed {\OURS} takes less time, only slightly slower than the random method. This is because the data used in {\OURS}, such as local momentum-based gradient, are already needed to be calculated in FL, so the additional computational overhead is not large. On the other hand, the mechanism based on multi-armed bandit fully utilizes historical information for exploration, so {\OURS} is relatively fast. In contrast, FedCor takes much more time to run for 200 rounds than other methods due to two factors. Firstly, FedCor has a warm-up stage where it collects information related to the quality of client data through several rounds of experiments. More importantly, FedCor needs to calculate the correlation between clients based on a Gaussian Process-based model in each round, which is time-consuming. In addition, the computational time required by FedCor for each round is proportional to the number of required clients $K$. The main reason for Pow-d's slightly slower performance compared to {\OURS} is the utilization of its post-selection strategy, which involves incorporating additional clients as candidates for model training before gradient aggregation. This approach increases the likelihood of stragglers and introduces overhead in synchronizing the parameters, thus impacting the overall performance.


\subsection{Effectiveness of Exploit-Explore}
\begin{figure}[ht]
\centering
\subfloat[Fixed $\alpha$]{     
\includegraphics[width=0.48\textwidth, trim=0 0 0 0, clip]{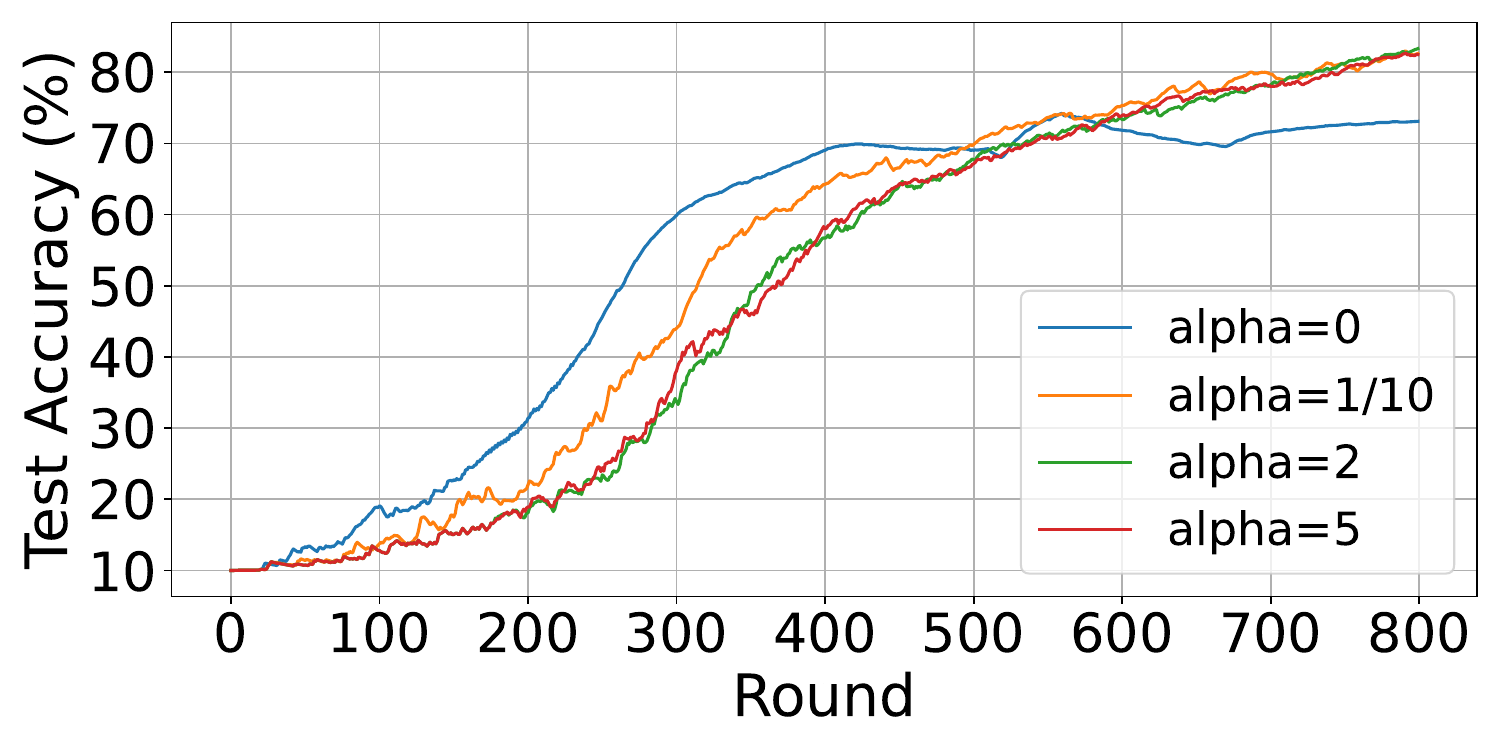}
\label{fig:alpha_const}
}
\vfill
\subfloat[Variable $\alpha$]{     
\includegraphics[width=0.48\textwidth, trim=0 0 0 0, clip]{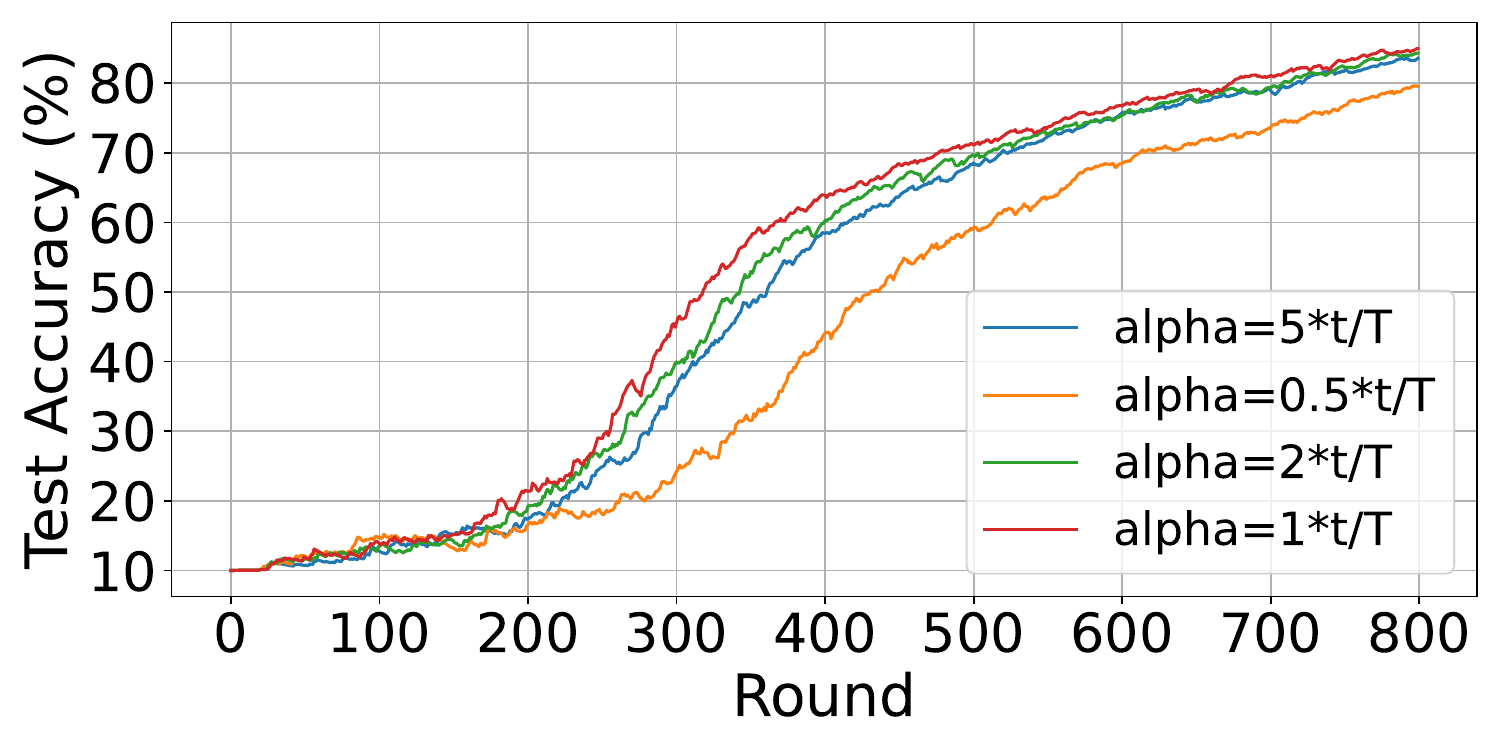}
\label{fig:alpha_linear}
}
\caption{Sensitivity of $\alpha$}
\label{Fig:alpha_fmnist}
\end{figure}
 In {\OURS}, we implemented an EE mechanism to select the best set of clients. The parameter $\alpha$ in Algorithm ref{alg:mgpcs2} determines the degree of exploration. A higher value of $\alpha$ encourages more exploration, while a lower value favours exploitation. We tested this mechanism on the FEMINST dataset by varying the value of $\alpha$. Our initial experiments involved using different fixed values of $\alpha$, as shown in Figure \ref{fig:alpha_const}. When $\alpha=0$, the model does not use the EE mechanism and selects the top $K$ clients with the highest GP values. In the first 500 rounds, not using EE mechanism yields good results, but performance plateaus after this point. However, when using EE mechanism, the results can still increase after 500 rounds, and the final result is nearly 10\% higher than that without EE. To accommodate different exploration requirements during the initial and later stages of training, {\OURS} employs a linear transformation of the parameter $\alpha$ based on the current round $t$ (Equation \ref{eq:alpha}). We experimented with different values of $\rho$ (0.5, 1, 2, and 5) in Equation \ref{eq:alpha}, and the results are presented in Figure \ref{fig:alpha_linear}. We found that a larger value of $\rho$ does not necessarily lead to better performance. In fact, the test accuracy is optimized when $rho$ is set to 1. Nevertheless, employing a linear transformation for $\alpha$ yields greater stability and higher test accuracy in the model's performance compared to using a fixed value.

%% file: 8Relatedwork.tex
\section{Related Work}


In Federated learning \cite{DBLP:journals/tkde/LiWWHWLLH23} \cite{DBLP:journals/csur/ChenZZZY24}, client selection algorithms are of paramount importance in determining which clients participate in the training process \cite{fu2023client}. Existing client selection methods can be broadly categorized into pre-selection and post-selection methods.

Federated learning pre-selection methods are designed to identify suitable clients for participation before the training process begins. These methods can be broadly classified into pre-selection based on client characteristics and pre-selection based on client performance or reliability. Pre-selection based on client characteristics involves considering factors such as computational capacity, bandwidth, or energy constraints of the clients \cite{xu2020client} \cite{yu2021toward}. By evaluating these characteristics, the algorithm can prioritize clients that are well-suited to perform the required computation and communication tasks efficiently. Additionally, client heterogeneity can be considered in the client selection process\cite{luo2022tackling}. Since heterogeneous statistical data can introduce biases during training, which can ultimately result in a degradation of accuracy in FL \cite{wang2020optimizing}. Pre-selection based on client performance or reliability could include historical accuracy, update quality, or communication stability \cite{deng2021auction} \cite{nishio2019client}. Assessing these metrics can help the algorithm favour clients that have demonstrated consistent performance or reliability in previous iterations, increasing the likelihood of obtaining high-quality updates from them.

Post-selection methods are used to choose clients based on their contributions or updates during the training process. These methods operate after the clients have finished their local training and generated updates. There are two common approaches for post-selection: client contribution-based methods and client diversity-based methods. Client contribution-based methods focus on identifying clients whose updates have the most positive impact on the global model. Factors such as the update's magnitude, its effect on the global model's performance \cite{nguyen2020fast}, or the novelty it brings \cite{lin2022contribution} are typically considered. By prioritizing clients with significant or innovative updates, these methods enhance the overall accuracy and convergence speed of the model. On the other hand, client diversity-based methods aim to select clients that represent various subsets of the client population. This diversity may be based on features, data distribution, or other relevant characteristics \cite{balakrishnan2022diverse} \cite{tang2022fedcor}. Including diverse clients helps mitigate bias and enables the global model to perform well across different client populations. This approach is particularly valuable in FL scenarios characterized by substantial heterogeneity among clients.

%% file: 9Conclusion.tex
\section{Conclusion}
In this paper, we propose GPFL, an efficient client selection framework for federated learning. GPFL introduces gradient projection (GP), a metric to evaluate client data quality. Clients calculate the gradient direction using momentum-based gradient with their local data, and the overall model's aggregated gradient direction is obtained through FedAvg. We assume this direction points to the minimum loss point, and GP is derived by projecting the client's gradient onto it. Drawing inspiration from the Upper Confidence Bound algorithm, we construct the Gradient Projection Confidence Bound using GP values. An exploration-exploitation mechanism is then employed to select an appropriate client set for federated learning. Experimental results demonstrate that {\OURS} achieves fast and accurate client selection on Non-IID data. However, this paper primarily focuses on client selection, while the model aggregation part utilizes the commonly used FedAvg algorithm. Future work will explore the design of a model aggregation algorithm that complements {\OURS} more effectively.